# Activity Recognition using Hierarchical Hidden Markov Models on Streaming Sensor Data

Parviz Asghari

Ambient Intelligence Research Lab.
Department of Computer Engineering
Amirkabir University of Technology
Tehran, Iran
parviz.asghari@aut.ac.ir

Ehsan Nazerfard

Ambient Intelligence Research Lab.
Department of Computer Engineering
Amirkabir University of Technology
Tehran, Iran
nazerfard@aut.ac.ir

*Abstract—* Activity recognition from sensor data deals with various challenges, such as overlapping activities, activity labeling, and activity detection. Although each challenge in the field of recognition has great importance, the most important one refers to online activity recognition. The present study tries to use online hierarchical hidden Markov model to detect an activity on the stream of sensor data which can predict the activity in the environment with any sensor event. The activity recognition samples were labeled by the statistical features such as the duration of activity. The results of our proposed method test on two different datasets of smart homes in the real world showed that one dataset has improved 4% and reached (59%) while the results reached 64.6% for the other data by using the best methods.

*Keywords—Activity Recognition, Hierarchical Hidden Markov models, Activity Segmentation, Sliding Window, Smart homes, Internet of things*

## I. INTRODUCTION

The human activity affects the person, society and the environment. Therefore, human activity recognition forms the basis of many research fields. Today, the advancement of sensor technology, as well as the evolution of machine learning methods, on the other hand, provides a good context for using sensors in various applications. One of these applications is smart environments that are used to monitor the behavior of a person living in the environment. For example, monitoring systems try to use activity recognition technology to deal with terrorist threats. Assisted living environments used activity recognition in order to help individuals to live independently. Other applications, for example, smart meeting rooms and smart hospitals, are also dependent on activity recognition. The activity recognition is also used in video game console and applications for patients' health and fitness on the smartphone. Smart environments have recently been used in many research laboratories throughout world; therefore, many research projects such as CASAS [1], MAVhome [2], PlaceLab [3], CARE [4], and AwareHome [5] have been deployed in this area.

Activity recognition refers to recognizing the daily activities of a person living in the environment. Activities of daily living (ADLs) is a terminology defined in healthcare to refer to people's daily self-care activities.

Activity recognition can be examined from different aspects:

### A. Type of Sensor

The first models were using visible features such as camera images to monitor individual behavior and environmental changes. The sensors data is a sequence of films or digital images. The sensor-based model uses a network of sensors to monitor the activity. In this approach, sensors can be attached to the person or placed in the environment.

### B. Activity model

In general, activity models are made by two methods. In the first method, a large set of data of each person's behavior are analyzed by using machine learning and data mining methods. In the second method, it was tried to obtain sufficient prior knowledge in the field of interests and personal tendencies by using knowledge engineering and management technologies. Activity recognition deals with various challenges, such as activity labeling, overlapping activities, sensors without useful information and activity extraction. Among these challenges, online activity recognition is identified as the main challenge in this area because the first step in providing real-world systems is to design systems which can recognize online user behavior.

In the present study, stream of data was made from a number of binary environmental sensors. Most methods for online activity recognition are based on the sliding window approach. These methods deal with the problem of optimization of the window size. The present study tries to discover the sequence pattern of the occurrence of the sensors at the beginning and ending of each activity to solve the window size problem, and also uses these patterns to identify the boundaries which means where the activity begins and ends on the stream of sensor data. As a result, it is not necessary to determine the window size, and the model automatically determines the proper size for activity recognition on the stream of data. When the beginning of the activity with the occurrence of each sensor is identified, the online activity will be predicted and this prediction continues until the corresponding completion pattern of the activity occurs. When the activity was identified, the activity recognition was labeled by the statistical features such as the duration of activity. The problem of sensors without information was also solved and the efficiency of the system performance increased. To implement this idea, the hierarchical Markov model was used. A set of hidden Markov models were responsible for recognizing of the beginning, ending, and type of activity patterns. This set was implemented as a hierarchical hidden Markov model. The results showed that the implemented algorithm was better than other available method or has the same performance.

This paper is organized as follows: In section 2, the previous studies were examined. Then, the proposed method and implementation details were described in Section 3. In section 4, various tests were performed to evaluate the efficiency of the model and compare the results of the proposed method with other methods. In the final section, conclusions and further studies were discussed.

## II. RELATED WORKS

The basis of activity recognition is to process of sequence events of the sensors and recognizes the corresponding activity. It is required to have a robust model for activity recognition due to the difference in the environmental structure and the sensors in our smart environments. The Diane proposed a robust model of the sensors to extract different features, such as the duration of the activity [6]. In the modeling, the problem was independent of the sensor environment and could be implemented for different users [7]. Since the accuracy of the model requires the data with proper labels, a method was proposed for the data labeling [8]. First, the data for activity recognition have already been segmented (i.e. the beginning and ending of each activity are identified) [9, 10]. In order to bring the activity recognition based systems closer to those of actual world, some methods were used for segmentation of the stream of the data [11, 12]. Activity labeling is one of the aspects and it is often ignored. Most researchers asked the residents to perform an activity and then mark the activity based on the activated sensors. This method was not practically possible for all people. Moreover, Szewcyzk et al. [8] presented an automated method to annotate the datasets. Other challenges for activity recognition were overlapping, and simultaneity of activities. Overlapping means that different classes have common sensors [13]. The next problem is those sensors which do not belong to any predefined classes; however, they usually include a large segment of the data set. Rashidi et al. [14] suggested that the corresponding new activity of these unprocessed sensors should be discovered, and added to predefined classes which improve the efficiency of the system. In addition to the challenges for activity modeling, machine learning methods were also discussed for activity recognition. Therefore, different methods including the ensemble method [15], the non-parametric method [16], the support vector machine method [9], as well as probabilistic models such as the hidden Markov model and the Markov random field have been used for activity recognition [17, 18, 19]. Each method has its own advantages and disadvantages. Probabilistic methods have high application due to high noise tolerance and the production of probability distributions over different classes [6].

Although each challenge has its own particular importance, online activity recognition is one of the most important challenges in the real world. Real world applications require systems which can recognize the user behavior immediately. Therefore, these methods should be able to process the stream of sensor data. Most methods for data stream processing are based on the sliding window approach [11, 20, 21, 22]. The sliding window approach, referred to as "windowing" is mainly based on the time or number of sensors. Initial methods considered the window size constant [18, 20]. Since different classes have different numbers of activated sensor, the sliding window size was proposed as a solution by many researchers [20, 22, 11, 23, 24]. Krishna et al. [20] mentioned the time dependence of the sensors as a criterion for dynamic window approach. They provided two window approaches based on the time and number of sensors. In the temporal sliding window approach, those sensors, which took place at a certain interval, were examined as one activity, while in the sensor-based method, first, those sensors which were continuously activated together, were recognized by using Mutual information criterion and then were classified in a window. The probabilistic approach was another proper idea which was provided by Fadi et al. [22]. In this approach, the window size was considered for each class. The initial size of the class was estimated according to the previous samples and was updated based on the sensor events. Yala et al. emphasized on the method [20] and the changes in computing mutual information and improved the system performance [21]. Kabir et al. [17] proposed a multi-stage method. In the first stage, the activities were clustered by a hidden Markov model and then the activities of each cluster were classified by using a separate hidden Markov model. Although the proposed methods have greatly solved the window size problem, there are still some problems such as the exact recognition of boundaries which reduces the efficiency of the system.

## III. THE PROPOSED METHOD

The problem of these methods was their inability in determining the window size precisely. Regarding the uncertainty about the window size and the approximate calculation of the window size, the recognition windows contain the events of sensors which do not belong to any predefined activities. The segmentation of the data without clear window size, and classification of the pieces as predefined classes and other class was proposed. Predefined classes included daily activities such as eating, bathing, sleeping, and so on, which were determined by medical specialists. Other classes are activities which are not included in this classification.

At first, the beginning and ending of the activities was tried to be recognized by identifying repetitive sensors in the beginning and ending of activities in order to segment the stream of data. After the recognition of the beginning of the activity, the corresponding activity will be predicted by each sensor event. When the activity ended and the corresponding piece was recognized, the pieces will be labeled by using the temporal features.

Figure 1 illustrates the steps of the proposed algorithm. In the activity discovery phase, when the stream of data entered, the sensors with information were first segmented and then classified based on the threshold of the recognized segments in the labeling phase. Each phase will be explained in the following sections.

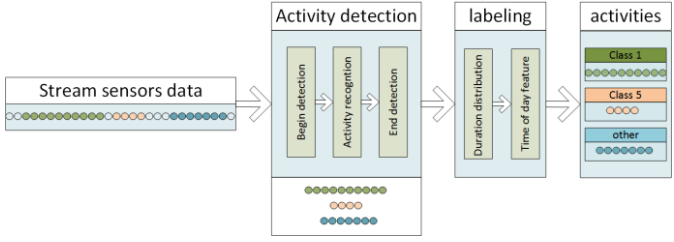

Figure 1 Steps of proposed method - phase one activity detection, phase 2 labeling

### A. Activity Discovery Phase

The first problem for activity recognition on streams of data is to recognize the occurrence of activities. The present study tries to examine the sensor event at the beginning and ending of each activity in order to recognize the start and end of it. As Figure 2 illustrates, sensors, which are active before and end of each activity, are usually constant. This is resulted from the fact that the person performs each activity in a specified location, and thus a specific set of sensors are activated at the beginning and the ending of the specified location.

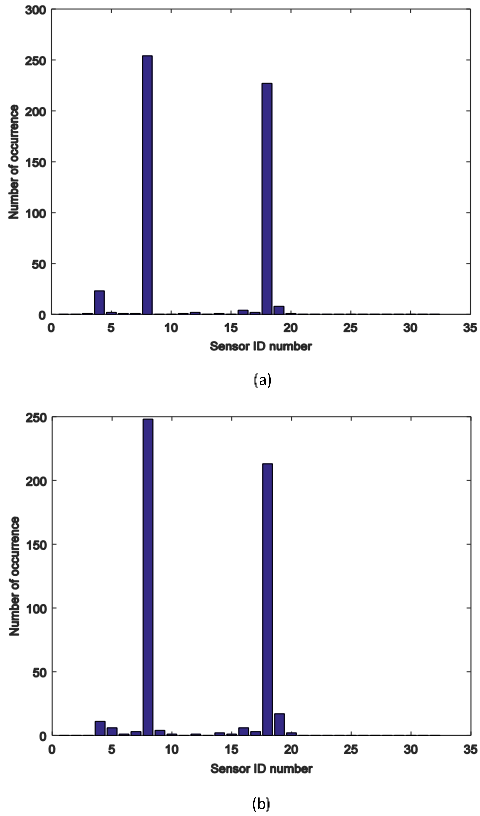

Figure 2 sensor frequency of personal hygiene activty. (a)before activity begins- (b) at the end of actvity

Some activities take place in a common place. For example, activities such as personal hygiene and bathing take place in one place. To distinguish these activities from each other, the sequence of sensor events is used. After the recognition of the beginning of the activity, the corresponding activity will be estimated by receiving each sensor. The more received sensors, the more accurate the recognized activity would be.

As it is mentioned earlier, the probabilistic models have an acceptable performance in dealing with noise data. In these models, the prior knowledge can be simply transferred to the models. Since the hidden Markov model has less computational load recursive neural networks than other similar methods, it is used as a common method for sequential data.

Suppose $X_t$ represent the hidden state vector and $Y_t$ represents vector observations. K is the number of possible hidden states for X. $X_t \in \{1,...,k\}$. Here, observations mean the sequence of sensor events. Equation 1 illustrates how to calculate the most probable class based on the observations.

$$P(X_t | y_{1:t}) = P(y_t | X_t)\left[\sum_{x_{t-1}} P(X_t | x_{t-1}) P(x_{t-1} | y_{1:t-1})\right] \quad (1)$$

Hidden Markov model will face difficulties in dealing with long sequences. Therefore, hierarchical hidden Markov model was selected. Hierarchical hidden Markov model is a closed version of the hidden Markov model which is appropriate for domains which have a hierarchical structure and multilevel dependencies on length and time.

Figure 3 illustrates the structure of the employed hierarchical hidden Markov model. In this network, dotted line vectors represent the vertical transition and solid line vectors represent horizontal transition. When abstract state is done, horizontal transition can occurs. In Figure 3, $X_1, X_2$, and $X_3$ represent abstract states. Each abstract state is a sub-HMM and produces sequence of observations. The lower nodes are production states that produce one observation.

The node $X_1$ is a sub-HMM which recognizes the beginning of the activity. Then, the control goes to node $X_2$. In this stage of the HHMM, the type of activity is recognized, and the corresponding activity will be predicted by sensor event of each sensor. After this stage, node $X_3$ recognizes the termination of the activity upon its completion. At the end, the control returns to the root node. In the output section, a part of sensors with information is identified which is shown as $S$ in the present study.

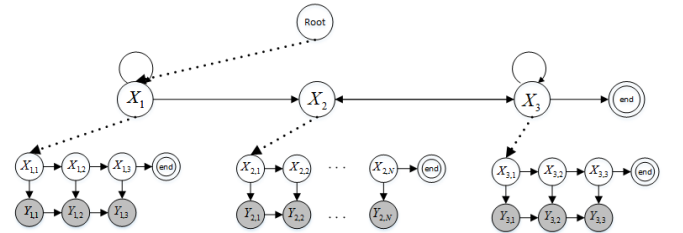

Figure 3 Implemented HHMM structure

### B. Activity Labeling Phase

In the dataset, undefined classes have not been labeled. Therefore, defined classes were used to identify these classes. The defined classes are denoted by "$C$". In fact, empirical probabilistic distribution of $C_k$ is extracted, and then the likelihood that $S$ belongs to $C_k$ is calculated. As equation 2 represents, if this probability exceeds a threshold, activity is considered as a class $C_k$, but if the probability is lower than the threshold, it is classified as the other class.

$$P(S \in C_k) = \begin{cases} 1 & f(t_s | T_{C_k}) \geq \alpha \\ 0 & f(t_s | T_{C_k}) < \alpha \end{cases} \quad (2)$$

Where $f(.)$, $\alpha$, $t_s$ and $T_{c_k}$ represent likelihood function, the threshold, duration of the activity segment $S$ and the duration empirical distribution of the class $C_k$, respectively. Also, $k$ denotes the class index, where $k \in \{1, 2, 3, ..., 11\}$.

## IV. DATASETS AND EXPERIMENTAL RESULTS

In order to evaluate the efficiency of the proposed model, the data of two intelligent homes were used, which will be described in the followings. Then, the evaluation criteria and the comparison methods were explained.

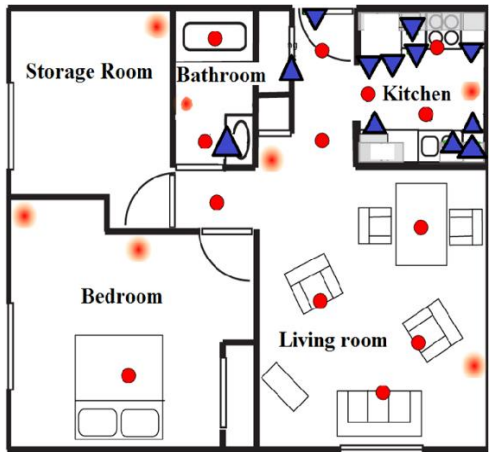

Figure 4 floor plan and sensor layout of home 1

Figure 4 illustrates the smart home architecture. This figure illustrates fixed motion sensors in red dots and door sensors in blue triangles. Table 1 illustrates a sample of this dataset.

Table 1 Characteristics of the smart homes

|  | Home 1 | Home 2 |
|---|---|---|
| # of motion sensors | 20 | 18 |
| # of door sensors | 12 | 12 |
| # of residents | 1 person | 1 person |
| # of sensor events collected | 371925 | 274920 |
| Timespan | 5 month | 4 month |

Table 1 illustrates the general specifications of home 1 and home 2 data sets and the used sensors. Activity classes include, personal hygiene, enter home, leave home, taking shower, cooking, relaxing on the couch, take medicine, eating, housekeeping, sleeping in bed, and bed to toilet transition.[1]

Of datasets, 70%, 10% and 20% were used for training, validation and testing, respectively.

---

[1] Results provided in the validation section correspond to the home 1 and home 2 datasets, available online at http://eecs.wsu.edu/~nazerfard/AIR/datasets/dataX.zip , where $X \in \{1, 2\}$.

Accuracy is one of the most commonly used evaluation criteria, and represents the proportion of the total number of positive and negative correct samples to the all samples (equation 3).

$$accuracy = \frac{tp + tn}{tp + tn + fp + fn} \quad (3)$$

### A. Estimatin the parameters

#### 1) The number of sensors preceeding an activity

To recognize the beginning of the activity, a number of different sensors has been investigated (Table 2).

Table 2 Estimating the number of preceeding sensors

|  | 2 | 3 | 4 | 5 | 6 |
|---|---|---|---|---|---|
| Home 1 | 78.7% | 97% | 92.9% | 93.2% | 93.1% |
| Home 2 | 84.3% | 96.4% | 94.2% | 91% | 91% |

Table 2 provides the accuracy of detecting the beginning of an activity. The results suggest that by considering three sensors, the best accuracy is archived.

#### 2) Threshold Parameter

In order to determine the appropriate threshold, the value of $\alpha$ has been changed from 0.02 to 0.10 and the optimum threshold was determined. Threshold values and different accuracy have been reported in Table 3.

Table 3 Estimating the alpha parameter

|  | 0.02 | 0.04 | 0.06 | 0.08 | 0.10 |
|---|---|---|---|---|---|
| Home 1 | 59.7% | 62.1% | 63% | 64.6% | 64.3% |
| Home 2 | 54% | 53% | 59% | 57% | 56.4% |

In table 4, the methods which were used for the comparison with the proposed model are described briefly.

Table 4 Notation and description of different approaches

| Notation | Description |
|---|---|
| Baseline | Baseline approach of fixed length Sliding window[20] |
| SWTW | Fixed length sensors windows with time based weighting sensors[20] |
| SWMI | Fixed length sensors windows with mutual information weighting sensors[21] |
| DW | Sensor windows where the window size determined dynamically[22] |

The results of the proposed method as well as the results of other methods are reported in Table 5. The numbers represent the efficiency of the algorithm based on the accuracy criterion.

Table 5 Accuracy Results

| Method | Home 1 | Home 2 |
|---|---|---|
| Proposed method | 64.6% | 59% |
| SWMI | 64% | 54% |
| SWTW | 62% | 45% |
| DW | 59% | 55% |
| Baseline | 58% | 48% |

As Table 5 illustrates, our proposed method performs better than the other approaches. The proposed method can also be used to represent an estimate of an activity at any given moment.

## V. CONCLUSIONS AND FUTURE WORK

Online activity recognition is one of the most important challenges in the area of smart environment research. Most of the previous methods for activity recognition had problem in determining the window size. The present study segmented the sensors for each event activity regardless of the window size by determining the beginning and ending of the activity. The results showed the efficiency of the proposed method compared to the available methods.

In general, the accuracy of the proposed methods for the activity recognition can exceed than a certain amount due to the uncertainty of human behavior. Therefore, it is practical to propose in the smart environment where activity is likely to occur. Because the training of activity recognition systems requires a large amount of data, it is recommended that transition learning and other people's information should be used to reduce the duration of collecting data and training system.